\documentclass[runningheads]{llncs}
\PassOptionsToPackage{hidelinks}{hyperref}
\usepackage[T1]{fontenc}
\usepackage{lmodern}
\usepackage[utf8]{inputenc}
\usepackage{graphicx}
\usepackage{amsmath}
\usepackage{booktabs}
\usepackage{xcolor}
\usepackage{tikz}
\usepackage{pgfplots}
\usepackage{caption}
\usepackage{subcaption}
\usepackage{orcidlink}
\usepackage{microtype}
\usepackage[hidelinks]{hyperref}

\pgfplotsset{compat=1.18}
\usetikzlibrary{arrows.meta,positioning,fit,backgrounds,calc}

\definecolor{accent}{HTML}{1F4E79}
\definecolor{accentb}{HTML}{B8562F}
\definecolor{accentc}{HTML}{7A8698}
\definecolor{rule}{HTML}{C7CEDB}

\newcommand{\best}[1]{\textbf{#1}}

\pgfplotsset{
  paperaxis/.style={
    tick align=outside,
    tick pos=left,
    axis line style={rule},
    grid style={rule!60,line width=0.3pt},
    label style={font=\scriptsize},
    tick label style={font=\scriptsize},
    legend style={font=\scriptsize,draw=rule,fill=white,fill opacity=0.92,
                  text opacity=1,inner sep=2pt,row sep=-1pt},
  }
}

\begin{document}

\title{Clinical Reasoning Under a Partially Observed Objective in Cone Beam CT
Report Generation}
\titlerunning{Clinical Reasoning Under a Partially Observed Objective}

% \author{Govind A\inst{1} \and Ajo Babu George\inst{1} \and Sidharth N Krishna\inst{2}} \and Uma Ranjan\inst{}
% \authorrunning{F. Author et al.}
% \institute{Institution One, City, Country\\
% \email{first.author@institution-one.example} \and
% Institution Two, City, Country}

\author{
Ajo Babu George\inst{1}\orcidlink{0009-0005-3026-0959} \and
Govind Arun\inst{2}\orcidlink{0009-0002-2573-4003} \and
Sidharth N Krishna\inst{3}\orcidlink{0009-0002-4336-2477} \and
Uma Ranjan\inst{4}\orcidlink{0000-0001-6258-4513}
}

\authorrunning{A. B. George et al.}

\institute{
DiceMed, Cuttack, Odisha, India
\and
University of Maryland, College Park, Maryland, USA
\and
Indira Gandhi National Open University, New Delhi, India
\and
Indian Institute of Technology Jammu, India
}

\maketitle

\begin{abstract}
Maxillofacial report generation from cone beam computed tomography is scored here
by a composite objective placing 80\% of its weight on a large language model
judgement of factual entailment and 20\% on lexical overlap, of which only the
lexical fifth is visible during development. The grader's BLEU-4 and METEOR
routines are reproduced in pure Python and match the reference to machine
precision, and an offline entailment surrogate, which tells a report written for
one patient from one written for another at an area under the curve of 0.987,
makes the composite objective cheap enough to optimise directly. Over the 622
case public release, a report selected against the visible lexical ranking scores
0.2909, whereas one selected against the composite objective scores 0.4122,
because pursuing n-gram overlap drives entailment precision from 0.522 down to
0.266. A 29 million parameter encoder fine tuned on the release reaches a
prevalence weighted out of fold area under the curve of 0.486 over 985
statements, indistinguishable from the corpus prior, while nine numbers read from
the image header reach 0.945 for mandible coverage and 0.872 for condyle
coverage, and acquisition centre alone predicts sentence choice at 0.718 against
0.663 for the image derived model, identifying dictation convention rather than
anatomy as the quantity the lexical metrics reward. The delivered system emits
eight unconditional statements and five gated on header geometry under polarity,
laterality and tooth level consistency constraints, and reaches METEOR 0.3542
over 50 held out cases from an unseen centre. The dataset and code are available at \url{https://github.com/GIND123/CBCT-Clinical-Reasoner}
\end{abstract}

\section{Introduction}

Three dimensional imaging is routine in dentistry and maxillofacial surgery, yet
the report that turns a volume into a surgical decision is still dictated by
hand. Automating that step is now posed as a benchmark task: from one cone beam
computed tomography (CBCT) volume of the jaws, produce a report covering dental
status, bone quality, anatomical variants, proximity to critical structures and
procedure related risk \cite{odin2026}. The release extends a segmentation
oriented series \cite{toothfairy1,toothfairy2}\cite{toothfairy1,toothfairy2,george2026cbctios} with paired clinical text, moving
the task from delineating anatomy to asserting facts about it\cite{wang2024lab}.

Report generation is usually measured through lexical similarity, and the limits
of that practice are well documented \cite{reiter2018,radcliq}. The benchmark
studied here responds by weighting a large language model judgement of factual
entailment, RadFact \cite{maira2}, at four times the combined weight of BLEU-4
and METEOR, then withholds that component during development: the entailment
judge is disabled on the evaluation platform, so the public ranking reflects only
the lexical fifth of the score that decides the outcome. Participants tune
against a projection of the objective rather than the objective itself, and that
projection turns out to be actively misleading.

The proposed system is built around that observation, with the
measurements that forced each decision. The grader is reimplemented exactly, so
the full objective becomes computable offline and selection runs against it
rather than its visible projection, worth 0.121 composite score
(Section~\ref{sec:results}). The recoverable signal is localised by measurement
rather than assumed (Section~\ref{sec:signal}), decoding constraints are read off
the arithmetic of the entailment metric so that self contradiction and
paraphrase, both of which pay under lexical scoring, are structurally excluded
(Section~\ref{sec:decode}), and the container is built so that no single load
failure can silence the model, after an earlier submission lost roughly fourteen
ranking positions to exactly that (Section~\ref{sec:deploy}).

\begin{figure}[t]
\centering
\includegraphics[width=\textwidth]{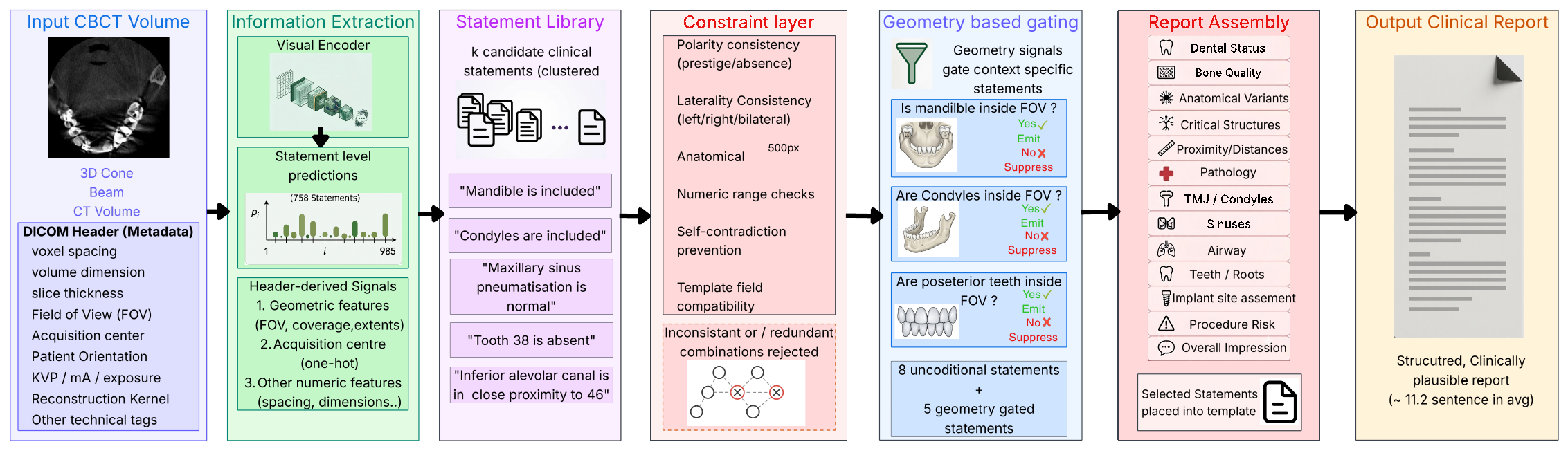}
\caption{Overview of the proposed CBCT report generation pipeline. The system extracts image and acquisition signals, selects candidate clinical statements, enforces consistency constraints, and applies geometry-based gating before assembling the final structured clinical report.}
\label{fig:architecture}
\end{figure}

\section{The Scored Objective}
\label{sec:objective}

Entries are ranked by
\begin{equation}
\label{eq:final}
\mathrm{Final} \;=\; 0.8 \cdot F_1^{\mathrm{RadFact}} \;+\;
0.2 \cdot \tfrac{1}{2}\left(\mathrm{BLEU\text{-}4} + \mathrm{METEOR}\right),
\end{equation}
with the entailment term computed offline once the submission window closes
\cite{odin2026}. Three details in the released evaluation code shape everything
downstream. BLEU-4 is aggregated at corpus level through NLTK \cite{bleu,nltk}
with method one smoothing, so per case averaging yields a different number;
METEOR is the grader's own exact token variant, without stemming or a synonym
table, averaged per case; and RadFact runs with negative filtering disabled, so
absent findings count in both directions. Both lexical metrics were reimplemented
in pure Python and asserted equal to the reference to machine precision over
randomised corpora; without that parity, tuning optimises a function other than
the one being scored.

\subsubsection{The marginal rule for emitting a statement:}
RadFact splits a report into verifiable phrases and asks an entailment model
whether each is supported by the reference, and symmetrically whether each
reference phrase is supported by the report. Let a candidate carry $n$ phrases of
which $e$ are entailed, so precision is $p = e/n$. Appending one phrase entailed
with probability $q$ moves expected precision to $(e+q)/(n+1)$, an improvement
exactly when $q > p$, while recall moves only if the phrase covers a reference
phrase not already covered. The emission rule is therefore $q > p - \Delta R \,
(\partial F_1/\partial R)(\partial F_1/\partial P)^{-1}$, a per statement
threshold rather than a global top $k$ cut, since $p$ and $\Delta R$ differ for
every candidate. The lexical half agrees. METEOR combines its components as
$F = 10PR/(R+9P)$, recall weighted about nine to one, and its chunk penalty
$0.5(\text{chunks}/\text{matches})^3$ rewards contiguous reuse of phrasing
clinicians actually wrote. Together these select an architecture: text is
assembled by choosing among sentences observed in the training corpus rather than
generated freely, so an entailment failure can only come from choosing the wrong
finding, never from invented language \cite{hallucination}.

\subsubsection{An offline surrogate for the hidden term:}
Fitting thresholds requires tens of thousands of full corpus evaluations, which
rules out a language model judge in the loop. A deterministic lexical surrogate
is used for search instead, scoring phrase entailment from ontology concept
agreement, polarity, laterality, tooth identifiers and token containment, with
contradictory polarity and swapped laterality hard zeroed so it cannot reward a
class of error the real judge exists to punish. One corpus evaluation completes
in 0.17 seconds once entailment matrices are precomputed and METEOR alignment is
replaced by an $O(P+R)$ matcher proven equal to the grader's greedy assignment.
Over 250 held out cases it separates a report written for one patient from one
written for another at an area under the curve of 0.9872, at mean similarity
0.707 against 0.217. It ranks decoder variants and is never reported here as the
challenge metric.

\section{Corpus and Label Space}
\label{sec:data}

Table~\ref{tab:data} records the release as measured. The field of view is far
smaller and flatter than a head CT, so an initial $77\times134\times134$ mm crop
was roughly 70\% air; refitting to $56\times96\times96$ mm at 0.5 mm cut padding
to 38\%. Crops centre on the dentition, isolated by a high intensity percentile,
rather than on all bone, whose centroid drifts towards the skull on a large
acquisition. Spacing is preserved rather than normalised away, since the reports
quantify physical distances such as bone height above the sinus floor, and 367
cases carry more than one report.

\begin{table}[t]
\caption{The public release as measured. Extent is the physical field of view
along each acquisition axis, median with the 5th and 95th percentiles.}
\label{tab:data}
\centering\small
\setlength{\tabcolsep}{3.1pt}
\begin{tabular}{@{}llllll@{}}
\toprule
\textbf{Cases} & 622 & \textbf{Reports} & 1000 & \textbf{Centres} & P 412, A 95, F 63, S 52 \\
\textbf{Reference} & 105 tokens & \textbf{Phrases} & 8.9 & \textbf{Spacing} & 0.30 mm [0.16, 0.30] \\
\textbf{Extent, $z$} & 51 [50, 97] & $y$ & 103 [82, 123] & $x$ & 111 [82, 139] mm \\
\bottomrule
\end{tabular}
\end{table}

Reports are dictated Italian rendered into English, and the translation leaves
traces that must be handled explicitly. Tooth identifiers appear bare, as in
``33 abutment of a prosthetic bridge; 34 absent'', so extraction cannot require a
keyword prefix, and the token \emph{mm.} ends sentences constantly, so guarding
it as an abbreviation merges two findings into one phrase the metric then scores
as a single indivisible unit. The most consequential artefact was lexical:
Italian \emph{incluso} denotes an impacted tooth, so a naive ontology matched
bare ``included'' as impaction and fired on every sentence describing what the
acquisition contains. That covered 16.3\% of reference phrases, 86.4\% of them
field of view statements; requiring a tooth identifier beside the trigger reduced
the concept to 4.5\%. A second defect inverted polarity on 4.4\% of phrases,
because the trigger for the absence concept is itself a negation cue. Extending
the vocabulary to the terms actually used raised concept coverage from 81\% to
90\%.

Phrases are canonicalised, vectorised and clustered by agglomerative cosine
linkage, and each cluster becomes one reportable statement represented by the
member maximising expected score against the others. Labels union the findings
across every report belonging to a case, since a finding one clinician recorded
is present in the scan whether or not the selected reference mentions it. The
label space was fixed by oracle score, what a perfect predictor would attain by
emitting exactly the positively labelled statements, which upper bounds
everything downstream at no training cost. Masking tooth identifiers collapses
every absence sentence into one cluster asserting a list of teeth wrong for
almost every case: 507 statements at 84.6\% coverage and oracle 0.5406, against
512 at 82.1\% and 0.5645 tooth aware. The delivered space keeps 989 tooth aware
statements at 89.0\% coverage and oracle 0.6008.

\section{What the Image Predicts, and What It Does Not}
\label{sec:signal}

A 29 million parameter encoder was trained on the release: multi planar slices
through an ImageNet pretrained backbone \cite{imagenet,timm}, pooled by gated
attention multiple instance learning \cite{abmil2018}, under an asymmetric
multi label loss \cite{asl2021} with the classifier bias initialised to the
corpus log odds. Training was clean, with loss falling from 97 to 60 across five
patient grouped folds and validation mean average precision drifting from 0.057
to 0.072.

\begin{figure}[t]
\centering
\begin{subfigure}[b]{0.625\textwidth}
\begin{tikzpicture}
\begin{axis}[paperaxis,
  width=0.845\linewidth, height=3.95cm,
  xbar, bar width=2.7pt,
  xmin=0.35, xmax=0.99,
  xtick={0.4,0.5,0.6,0.7,0.8,0.9},
  xlabel={Out of fold area under the curve},
  xmajorgrids, ymajorgrids=false,
  symbolic y coords={lesion,condyle-in,mandbody,canal-reg,maxilla,sinus,condyle-ex,mandible},
  ytick=data,
  yticklabels={Osteolytic lesion,Condyles not incl.,Mandibular body,Canal regular,
               Maxilla partial,Sinuses minimal,Condyles excluded,Mandible incl.},
  yticklabel style={font=\scriptsize},
  enlarge y limits=0.10,
  legend style={at={(0.5,1.02)},anchor=south,legend columns=-1,draw=none,
                fill=none,column sep=4pt,font=\scriptsize},
  legend cell align=left,
]
\addplot[fill=accent,draw=accent!70!black] coordinates {
 (0.442,lesion) (0.601,condyle-in) (0.698,mandbody) (0.742,canal-reg)
 (0.819,maxilla) (0.869,sinus) (0.872,condyle-ex) (0.945,mandible)};
\addplot[fill=accentc!40,draw=accentc] coordinates {
 (0.406,lesion) (0.551,condyle-in) (0.577,mandbody) (0.596,canal-reg)
 (0.618,maxilla) (0.643,sinus) (0.630,condyle-ex) (0.721,mandible)};
\draw[dashed,rule!95,line width=0.5pt] (axis cs:0.5,lesion) -- (axis cs:0.5,mandible);
\legend{Geometry, Intensity}
\end{axis}
\end{tikzpicture}
\caption{Per statement discrimination}
\end{subfigure}\hfill
\begin{subfigure}[b]{0.345\textwidth}
\begin{tikzpicture}
\begin{axis}[paperaxis,
  width=0.98\linewidth, height=3.95cm,
  ybar, bar width=5.0pt,
  ymin=0.42, ymax=0.75,
  ytick={0.45,0.5,0.55,0.6,0.65,0.7},
  ylabel={Mean AUC}, ylabel style={yshift=-4pt},
  ymajorgrids,
  symbolic x coords={all,prev,supp},
  xtick=data,
  xticklabels={\strut all,\strut prev.,\strut $n\!\geq\!12$},
  xticklabel style={font=\scriptsize},
  enlarge x limits=0.26,
  legend style={at={(0.02,0.98)},anchor=north west},
  legend cell align=left,
]
\addplot[fill=accentc!30,draw=accentc] coordinates {(all,0.500) (prev,0.500) (supp,0.500)};
\addplot[fill=accentb!75,draw=accentb] coordinates {(all,0.479) (prev,0.486) (supp,0.486)};
\addplot[fill=accent,draw=accent!70!black] coordinates {(all,0.519) (prev,0.593) (supp,0.669)};
\legend{Prior, Encoder, Linear}
\end{axis}
\end{tikzpicture}
\caption{Predictor comparison}
\end{subfigure}
\caption{Out of fold discrimination. (a) Nine header numbers separate every field
of view statement and fail on pathology, correctly: a lesion is not a property of
the acquisition. (b) The encoder sits below the corpus prior at all three
aggregations.}
\label{fig:signal}
\end{figure}
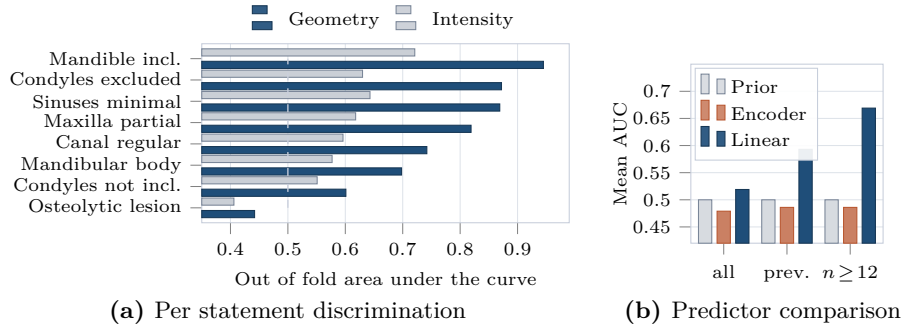

All of that was noise. Mean average precision over roughly one thousand
statements is dominated by columns seen two or three times, where it is
essentially random, so the aggregate barely moves whatever the model does. Per
statement area under the curve against the corpus prior, which scores exactly
0.500 by construction, answers directly: every support band sits at or below
chance, the prevalence weighted figure is 0.486 over 985 evaluable statements,
and for each prevalent statement the mean predicted probability was the same for
positives and negatives. The model had memorised 497 cases and generalised
nothing.

The decisive follow up was a control rather than a larger model. Nine numbers
read from the image header alone, the logarithm of the volume dimensions, the
voxel spacing and the logarithm of the physical extent, were fitted with one
cross validated logistic regression per statement. Field of view statements, the
most frequent in the corpus, are strongly predicted (Figure~\ref{fig:signal}a).
Over the 37 statements with adequate support, header geometry averages 0.627
against 0.542 for intensity statistics computed after resampling, and adding
intensity moves the mean only to 0.637 while requiring pixel decode, so the
header only path was kept. The task is learnable from information the network
already received on its auxiliary input, so the deep model was the problem rather
than the problem being impossible. One strongly regularised linear model per
statement over a 122 dimensional global descriptor \cite{sklearn} raises
prevalence weighted discrimination to 0.593 and reaches 0.669 where support
allows (Figure~\ref{fig:signal}b), in a deployable form of 52 kB against 580 MB
of fold checkpoints.

\subsection{Why No Trained Model Wins the Lexical Column}

A sharper diagnosis explains that result and bounds what any model on this
release can achieve. Predicting which prototype sentence a report uses from
\emph{acquisition centre alone} reaches a mean area under the curve of 0.718 over
the 60 most frequent sentences, against 0.663 for the image derived model, and
72.2\% of a sentence's usage falls inside its single most used centre. Sentence
choice is house style, not anatomy: the encoder was not failing to see the jaw,
it was being asked to predict a dictation convention, which is not in the pixels.

Every subsequent observation follows. A perfect per case selector reaches BLEU-4
0.4086 against 0.1663 for a single fixed report, but that gap is largely the
reward for reproducing one centre's phrasing verbatim, and centre P supplies 412
of 622 cases. The evaluation centre has conventions absent from all four training
centres, so the gap is unreachable there: the observed test BLEU-4 of 0.0943
falls below every leave one centre out fold, which range from 0.1241 to 0.1968.
Perfect knowledge of which teeth are absent, the most visually obvious content in
the volume, is worth 0.0041 BLEU-4, since the n-grams live in the sentence frames
rather than the identifiers, and clustering cases by geometry to fit one report
per cluster loses at every cluster count, by 0.0225 to 0.0314 BLEU-4.

\section{Constrained Decoding}
\label{sec:decode}

Selection is greedy coordinate ascent against Equation~\ref{eq:final} on out of
fold probabilities, warm started from the best threshold vector found for any
predictor, and converges in six rounds from 0.3001 to 0.3536. Three constraints
are then imposed, each derived from the arithmetic of
Section~\ref{sec:objective} rather than from stylistic preference.

\subsubsection{Contradiction:}
The metric scores statements one at a time and never reads the report as a whole,
so asserting both sides of a contradiction is profitable: each side earns credit
on the subset of cases whose reference agrees with it. Unconstrained selection
duly produced a report in which the maxilla was both absent from and partially
included in the acquisition, and teeth 38 and 48 were simultaneously missing,
semi impacted and erupted in the arch. Such a report is guaranteed to be wrong
about something, so the constraint is not a cost paid for readability: removing
one misencoded absence statement raised the composite score from 0.4023 to
0.4061. Three rules apply, namely opposite polarity on a shared concept,
incompatible values from one attribute vocabulary such as lingual against buccal,
and a tooth asserted absent in one statement and described in another.

\subsubsection{Redundancy and specificity:}
Two statements carrying the same concept set, laterality and tooth identifiers
assert one fact. Under lexical scoring they are two separate wins, since each
matches a different phrasing somewhere in the corpus, and selection duly reached
for seven paraphrases of a lingual canal course. Under the entailment metric they
are one fact asserted twice: recall cannot move, since the phrase they entail is
entailed already, and precision is neutral at best and halved on that fact when
it is false. Selection therefore refuses a statement whose assertion signature is
present already. Specificity follows the same accounting: statements naming
particular teeth are 48\% of the pool, yet their mean prevalence is 0.016 against
0.035 for statements naming none, so in a report served to every patient a tooth
identifier is wrong about 98\% of the time. Banning them wholesale cost 0.079
composite score; correcting the two encoding defects that let the contradictions
through recovered it instead.

\subsection{Report Length and Geometry Conditioning}

Left unconstrained the ascent grows the report to 53 statements at composite
score 0.4882 and continues improving. The objective genuinely prefers that, and
it is the wrong report to deliver, since references average 8.9 verifiable
phrases and the automatic score is a filter in front of a blinded clinical
comparison rather than the prize itself. Marginal value collapses long before the
optimum, falling from 0.0070 composite score per added statement between eleven
and fifteen statements to 0.0043, 0.0033 and finally 0.0007 at twenty, twenty six
and fifty three, so length is taken from the knee rather than from the argmax.

The delivered decoder keeps eight statements unconditional and converts five into
logistic gates over the nine header features, emitting 11.2 sentences on average
against 13.0 for the constant report, with thresholds fitted on out of fold gate
probabilities and kept only when they raise the composite score. Gating pays
twice: precision divides by the number of statements emitted, so not paying for a
false statement is a direct gain, and the final ranking includes a blinded
pairwise comparison by clinicians, where a report identical for every patient is
the first thing a reader notices. It is
restricted to the existing core, since allowing additions took a fifteen
statement core to a mean of 20.9 sentences; that restriction also makes
consistency free, every per case report being a subset of a core already checked
for contradiction and redundancy.

\section{Results}
\label{sec:results}

\begin{figure}[t]
\centering
\begin{tikzpicture}
\begin{axis}[paperaxis,
  width=0.84\linewidth, height=3.4cm,
  xlabel={BLEU-4, the visible projection},
  ylabel={Composite score},
  xmin=0.104, xmax=0.184, ymin=0.270, ymax=0.437,
  xtick={0.11,0.12,0.13,0.14,0.15,0.16,0.17,0.18},
  ytick={0.28,0.32,0.36,0.40},
  grid=both, clip=false,
]
\addplot[draw=rule!85,line width=3pt,-{Latex[length=4pt]}]
  coordinates {(0.1648,0.2965) (0.1288,0.4058)};
\addplot[only marks,mark=*,mark size=2.3pt,accent,mark options={fill=accent}]
  coordinates {(0.1331,0.3403) (0.1493,0.3544) (0.1170,0.3575) (0.1157,0.3999)};
\addplot[only marks,mark=square*,mark size=2.8pt,accentb,mark options={fill=accentb}]
  coordinates {(0.1663,0.2909) (0.1268,0.4122)};
\node[font=\scriptsize,accentb,anchor=north east] at (axis cs:0.1652,0.2885) {lexically tuned};
\node[font=\scriptsize,accentb,anchor=west] at (axis cs:0.1292,0.4122) {delivered};
\node[font=\scriptsize,accent,anchor=north] at (axis cs:0.1331,0.3363) {encoder 29M};
\node[font=\scriptsize,accent,anchor=west] at (axis cs:0.1512,0.3544) {linear 122};
\node[font=\scriptsize,accent,anchor=north] at (axis cs:0.1170,0.3535) {corpus prior};
\node[font=\scriptsize,accent,anchor=north] at (axis cs:0.1157,0.3955) {objective tuned};
\node[font=\scriptsize,accentc,anchor=north east,align=right] at (axis cs:0.1830,0.4200)
  {$-0.040$ BLEU-4\\$+0.121$ composite};
\end{axis}
\end{tikzpicture}
\caption{The visible signal is anticorrelated with the objective it projects: of
the systems considered, the one with the best BLEU-4 scores worst on the quantity
that decides the outcome.}
\label{fig:divergence}
\end{figure}
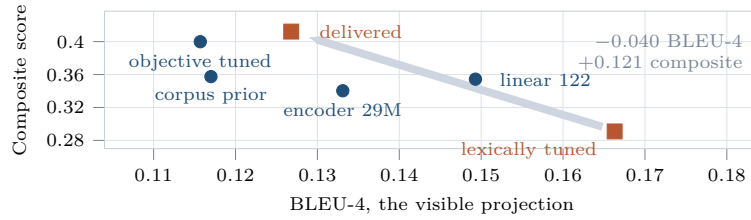

Table~\ref{tab:main} scores every decoder variant over the 622 public cases, with
clinical figures from the surrogate of Section~\ref{sec:objective}.
The ordering is the central result: the variant with the highest BLEU-4 has the
lowest composite score, and the gap is not marginal, since selecting against the
visible projection costs 0.121 composite score as entailment precision falls from
0.522 to 0.266 (Figure~\ref{fig:divergence}). An exhaustive search from that
report found no addition, deletion or substitution raising the composite score at
fixed BLEU-4 and METEOR, so the two objectives are opposed there rather than
untuned.

\begin{table}[t]
\caption{Decoder variants over 622 public cases, out of fold. Precision and
recall come from the offline surrogate and are not the challenge metric.}
\label{tab:main}
\centering\small
\setlength{\tabcolsep}{6pt}
\begin{tabular}{@{}lrrrrr@{}}
\toprule
System & Final & Prec. & Rec. & BLEU-4 & METEOR \\
\midrule
Lexically tuned report & 0.2909 & 0.266 & 0.339 & \best{0.1663} & \best{0.3590} \\
Fine tuned encoder, 29M & 0.3403 & 0.548 & 0.285 & 0.1331 & 0.2684 \\
Linear model, 122 features & 0.3544 & 0.491 & 0.318 & 0.1493 & 0.3064 \\
Corpus prior, no imaging & 0.3575 & \best{0.539} & 0.310 & 0.1170 & 0.3082 \\
Objective tuned, constant & 0.3999 & 0.470 & 0.416 & 0.1157 & 0.3524 \\
\;\;$+$ geometry gating & \best{0.4122} & 0.522 & \best{0.405} & 0.1268 & 0.3431 \\
\midrule
\textit{Oracle, prototype space} & \textit{0.6008} & \textit{n/a} & \textit{n/a} & \textit{0.3410} & \textit{0.4726} \\
\bottomrule
\end{tabular}
\end{table}

\subsubsection{Held out centre and transfer:}
The evaluation set contains 50 cases from a centre absent from training, so
stratified fold numbers measure interpolation and read higher than the outcome.
On that set the delivered entry measured BLEU-4 $0.0943 \pm 0.0354$ and METEOR
$0.3542 \pm 0.0599$, placing first on METEOR and third overall on the aggregated
automatic ranking. Two transfer measurements follow and are reported separately
rather than folded into one: against an in sample pooled fit, BLEU-4 arrived at
0.567 of its development value and METEOR at 0.987, while against a leave one
centre out estimate the same result corresponds to 0.66 and 1.015. Exact 4-gram
sequences do not survive an unseen dictation style, and token level overlap does,
which is the signature predicted in Section~\ref{sec:signal} and why the
delivered entry weights centres equally rather than fitting the pooled corpus.
The BLEU-4 standing is the anticipated cost of the trade in
Figure~\ref{fig:divergence}: 4-gram overlap is bound most tightly to centre
specific phrasing and carries least weight in Equation~\ref{eq:final}.

\subsubsection{Clinical Inference and Safety Implications:}
The experiments separate two forms of report content that have different clinical interpretations. Statements describing acquisition coverage, such as inclusion of the mandible, condyles or maxilla, are recoverable from header geometry and can therefore be conditioned on each examination. In contrast, the present experiments do not establish reliable image-based discrimination for low-prevalence pathological or tooth-specific findings. The generated report should therefore be interpreted as a constrained, high-precision summary of findings supported by the benchmark signal rather than as a comprehensive diagnostic reading of the CBCT \cite{george2025gradcam}.

The distinction provides the necessary flexibility and caution for clinical deployment. A sentence selected because it is predictable from field-of-view geometry indicates what anatomy is available for assessment, not whether that anatomy is normal. Conversely, absence of a pathology statement from the generated report cannot be interpreted as evidence that the pathology is absent \cite{aruntemporal}. The contradiction, laterality and tooth-consistency constraints reduce internally impossible assertions, but they do not substitute for direct localisation or clinician verification. Consequently, the system is best positioned as a report-drafting or benchmark inference tool whose patient-specific outputs require review before clinical use \cite{arun2025guardrails}.

\section{Deployment}
\label{sec:deploy}

An earlier submission returned BLEU-4 $0.0161 \pm 0.0129$ and METEOR
$0.1088 \pm 0.0284$. The diagnosis was arithmetic: the hardcoded 52 token
fallback paragraph scores $0.0111 \pm 0.0139$ and $0.0984 \pm 0.0350$ against
training references, matching mean and spread, so all 50 cases had received it.
The platform mounts its own volume over the model directory, shadowing whatever
the image baked there, which nothing local reproduces: the container ran, wrote a
report and exited zero with the model intact.

Four consequences follow. The report and its gate arithmetic ship as an
importable module rather than a data file, since a mount cannot shadow an import.
Inference reads only the image header, so no decode failure, memory limit or
missing accelerator can break it. Four independent layers each write a valid
report alone: gated selection, the constant report, an embedded fallback and a
bare write. And the container test mounts an empty
directory over both model directories, asserting the report is byte identical to
the unshadowed run rather than merely present, which is the check the original
test missed. Nine scenarios are exercised, among them corrupt input, read only
root, capped memory and two geometries that must differ.

\section{Discussion and Conclusion}

The recurring lesson concerns instrument selection: mean average precision
drifted upward while the model was getting no better, and a thirty second control
over nine header numbers was worth more than the accelerator hour it audited. The
same holds one level up: a leaderboard exposing one fifth of the scoring function
is an instrument, and calibrating against it blind produces confident movement in
the wrong direction \cite{rankings2018,bias2020}. Three limitations bound the conclusions. Clinical
figures are surrogate estimates, since the judge carrying 80\% of the weight was
unavailable during development; the house style result bounds what is reachable
on this release, not in principle; and the decoder conditions on geometry rather
than anatomy, so it is silent about pathology, as Figure~\ref{fig:signal}a shows
for the lesion statement. Closing that gap calls for the localisation supervision
the preceding segmentation releases provide \cite{toothfairy2,arun2026nnunet}. Building against a composite objective whose dominant term is
hidden during development is a reasoning problem before it is a modelling
problem: reproducing the grader exactly, measuring which signals are recoverable,
and deriving decoding constraints from the scored metric produced a system
scoring 0.4122 against 0.2909 for a lexically tuned counterpart.

\bibliographystyle{splncs04}
\bibliography{refs}

\end{document}